\documentclass[10pt,twocolumn,letterpaper]{article}

\usepackage{iccv}

\usepackage{times}
\usepackage{epsfig}
\usepackage{graphicx}
\usepackage{amsmath}
\usepackage{amssymb}

\usepackage{amsmath,amsfonts,bm}

\def\eqref#1{equation~\ref{#1}}

\def\1{\bm{1}}

\def\va{{\bm{a}}}

\def\vh{{\bm{h}}}

\def\vx{{\bm{x}}}
\def\vy{{\bm{y}}}
\def\vz{{\bm{z}}}

\DeclareMathAlphabet{\mathsfit}{\encodingdefault}{\sfdefault}{m}{sl}
\SetMathAlphabet{\mathsfit}{bold}{\encodingdefault}{\sfdefault}{bx}{n}

\graphicspath{{files/graphics/}}
\DeclareGraphicsExtensions{.pdf,.jpeg,.png}
\usepackage{import}
\usepackage{algorithm}
\usepackage{algpseudocode}
\usepackage{caption}
\usepackage{subcaption}
\usepackage{booktabs}
\usepackage{multirow}
\usepackage{mathtools}
\usepackage{color, colortbl}
\usepackage{multirow, float, booktabs, boldline, hhline}
\usepackage{xcolor}

\usepackage[pagebackref=true,breaklinks=true,letterpaper=true,colorlinks,bookmarks=false]{hyperref}

\def\iccvPaperID{1903}

\iccvfinalcopy
\ificcvfinal
\fi

\begin{document}

\title{Semantics Disentangling for Generalized Zero-Shot Learning}

\author{Zhi Chen$^1$ \quad  Yadan Luo$^1$ \quad  Ruihong Qiu$^1$ \quad  Sen Wang$^1$ \quad Zi Huang$^1$ \quad  Jingjing Li$^2$ \quad  Zheng Zhang$^3$ \\ 
$^1$ The University of Queensland, Australia  \\ $^2$University of Electronic Science and Technology of China   $^3$Harbin Institute of Technology, Shenzhen  \\ 
\{zhi.chen,y.luo,r.qiu,sen.wang\}@uq.edu.au \\ huang@itee.uq.edu.au,  lijin117@yeah.net, darrenzz219@gmail.com}

\maketitle
\ificcvfinal\thispagestyle{empty}\fi

\begin{abstract}
    Generalized zero-shot learning (GZSL) aims to classify samples under the assumption that some classes are not observable during training. To bridge the gap between the seen and unseen classes, most GZSL methods attempt to associate the visual features of seen classes with attributes or to generate unseen samples directly. Nevertheless, the visual features used in the prior approaches do not necessarily encode semantically related information that the shared attributes refer to, which degrades the model generalization to unseen classes. To address this issue, in this paper, we propose a novel semantics disentangling framework for the generalized zero-shot learning task (SDGZSL), where the visual features of unseen classes are firstly estimated by a conditional VAE and then factorized into semantic-consistent and semantic-unrelated latent vectors. In particular, a total correlation penalty is applied to guarantee the independence between the two factorized representations, and the semantic consistency of which is measured by the derived relation network. Extensive experiments conducted on four GZSL benchmark datasets have evidenced that the semantic-consistent features disentangled by the proposed SDGZSL are more generalizable in tasks of canonical and generalized zero-shot learning. Our source code is available at \url{https://github.com/uqzhichen/SDGZSL}. 
\end{abstract}

\section{Introduction}
\label{sec:introduction}
Human beings have a remarkable ability to learn new notions based on prior experience without seeing them in advance. For example, given the clues that zebras appear like horses yet with black-and-white stripes, one can quickly recognize a zebra if he/she has seen horses before. Nevertheless, unlike humans, supervised machine learning algorithms can only classify samples belonging to the classes that have already appeared during the training phase, and they are not able to handle samples from previously unseen categories. This challenge motivates the study of generalizing models to the unseen classes by transferring knowledge from intermediate semantics (\textit{e.g.,} attributes), which typically refers to zero-shot learning (ZSL).
\begin{figure}
    \centering
    \includegraphics[width=78mm]{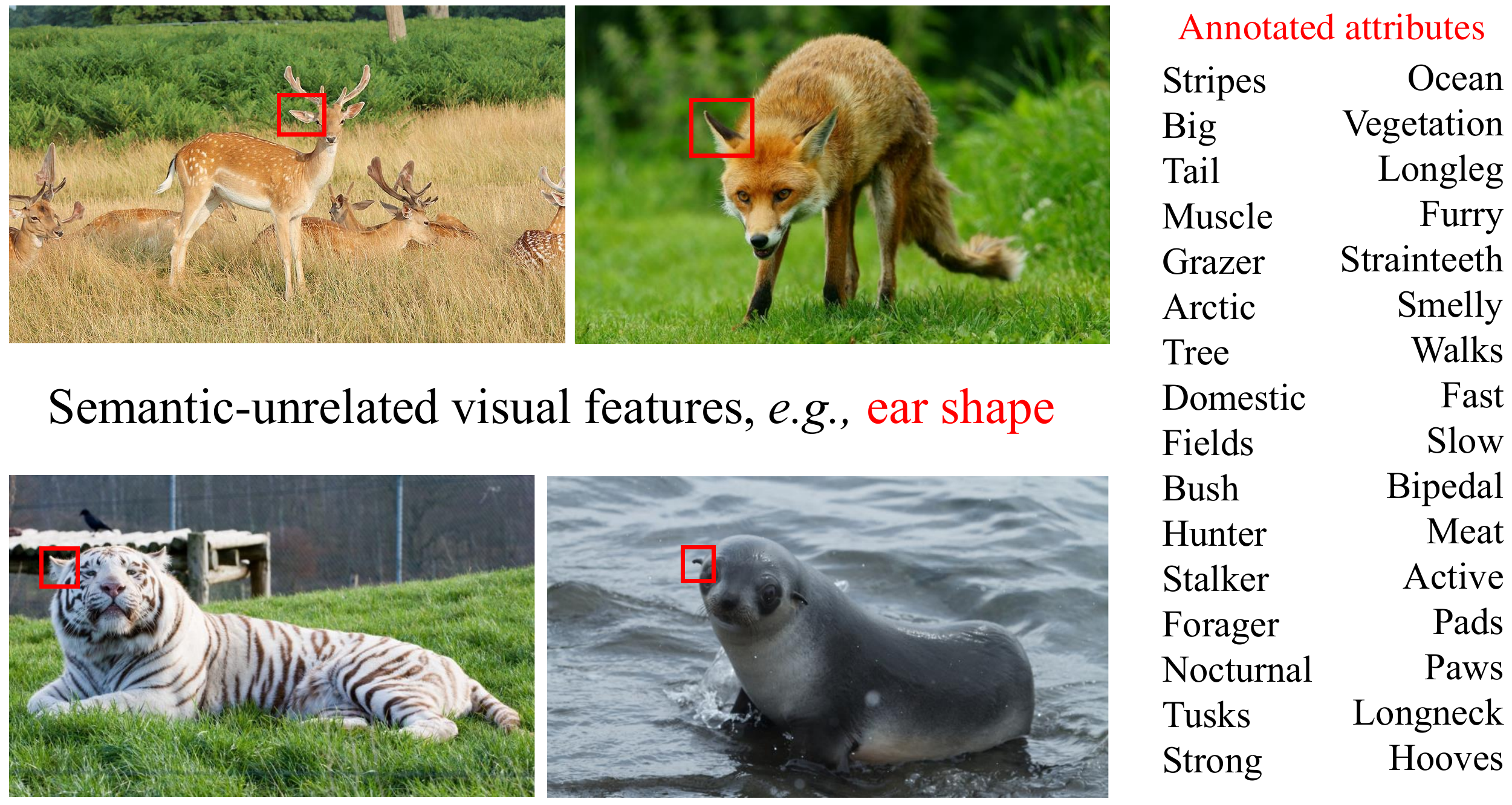}
    \caption{An illustration of the visual features (red boxes) that are not associated with the annotated attributes. Learning from such visual features that are semantically unrelated may jeopardize the model generalization to unseen classes.}
    \label{att}
    \vspace{-10pt}
\end{figure}intro

Particularly, the core idea of ZSL \cite{lampert2013attribute,xian2016latent, akata2015label, kodirov2017semantic} lies in learning to map features between the semantic space and visual space, thereby closing the gap between the seen and unseen classes. While effective, conventional ZSL techniques are built upon the assumption that the test set only contains samples from the unseen classes, which can be easily violated in practice. Hence, it is more reasonable to consider a new protocol called generalized zero-shot learning (GZSL), where seen and unseen images are both to be identified.

Existing GZSL techniques can be roughly grouped into two types: embedding-based \cite{frome2013devise,xian2016latent,liu2019attribute,liu2018generalized,jiang2019transferable} and generative-based \cite{zhu2018generative, xian2018feature, schonfeld2019generalized, narayan2020latent, keshari2020generalized} approaches. The former group learns a projection or an embedding function to associate the visual features of seen classes with the respective semantic vectors, while the latter one learns a visual generator for the unseen classes based on the seen points and semantic representations of both classes. However, most GZSL approaches directly leverage the visual features extracted from the pre-trained deep models, such as ResNet101 \cite{he2016deep} pre-trained on ImageNet, which are not tailored for ZSL tasks. In \cite{tong2019hierarchical}, it is observed that not all the dimensions of the extracted visual features are semantically related to the pre-defined attributes, which triggers the bias on learning semantic-visual alignment and causes negative transfer to unseen classes. Given an example from the AWA dataset shown in Figure \ref{att}, despite the features of animals' ears are visually salient for discriminating image samples, it is ignored in the manually annotated attributes. When generalizing to unseen classes such as cats, it is easy for them to be misclassified as tigers because the visual features corresponding to the concepts ``\textit{Big, Strong, Muscle}'' are not highlighted. From this case, we believe that GZSL will benefit from using the visual features that can consistently align with the respective semantic attributes. We define this type of visual features as the semantic-consistent features, which are agnostic to both seen and unseen classes. In contrast, those visual features that are irrelevant to manually annotated attributes are defined as semantic-unrelated. 

To unravel semantic-consistent and semantic-unrelated features from the original visual spaces, we present a novel framework, namely Semantics Disentangling for Generalized Zero-Shot Learning (SDGZSL), as shown in Figure 2.
 Specifically, we disentangle the underlying information of the extracted visual features into two disjoint latent vectors $\vh_s$ and $\vh_n$. They are learned in an encoder-decoder architecture with a relation module and a total correlation penalty. The encoder network projects the original visual features to $\vh_s$ and $\vh_n$. To make $\vh_s$ consistent with the semantic embeddings, the relation module calculates a compatibility score between $\vh_s$ and semantic information to guide the learning of $\vh_s$. We further apply the total correlation penalty to enforce the independence between $\vh_s$ and $\vh_n$.
Afterward, we reconstruct the original visual features $\bar{\vx}$ from the two latent representations. This reconstruction objective ensures the two latent representations to cover both semantic-consistent and semantic-unrelated information. 
The disentangling modules are incorporated into a conditional variational autoencoder and trained in an end-to-end manner.
The proposed framework is evaluated on various GZSL benchmarks and achieves better performance compared to the state-of-the-art methods. 
The main contributions of this work are summarized as follows:
\begin{itemize}
    \item
 We propose a novel feature disentangling framework, namely Semantic Disentangling for Generalized Zero-Shot Learning (SDGZSL), to disentangle the underlying information of visual features into two latent representations that are semantic-consistent and semantic-unrelated, respectively. Exploiting the semantic-consistent representations can substantially increase the performance in GZSL comparing to directly using entangled visual features that are extracted from the pre-trained CNN models.
 \vspace{-5pt}
 \item To facilitate the feature disentanglement of the semantic-consistent and semantic-unrelated representations, by introducing a total correlation penalty in our framework we arrive at a more accurate characterization of the semantically annotated features.
 \vspace{-5pt}
\item Extensive experiments conducted on four benchmark datasets evidence that the proposed method performs better than the state-of-the-art methods. 
\end{itemize}

\section{Related Work}

Recent state-of-the-art approaches for GZSL using generative models have achieved promising performance. Generative models can synthesize an unlimited number of visual features from side information of the unseen classes, \textit{e.g.,} manually annotated attributes. With these synthesized features, ZSL problems become a relatively straightforward supervised classification task. The two most commonly used generative models are generative adversarial networks (GANs) \cite{goodfellow2014generative} and variational autoencoders (VAEs) \cite{VAE}. Often, both models are jointly used to form generative architectures for ZSL tasks. f-CLSWGAN \cite{xian2018feature} leverages Wasserstein GANs (WGAN) \cite{arjovsky2017wasserstein} to synthesize vivid visual features.
CADA-VAE \cite{schonfeld2019generalized} leverages two aligned variational autoencoders to learn the shared latent representations between different modalities.
SE-ZSL \cite{kumar2018generalized} adopts an autoencoder followed by an attribute regressor to train a model with three alignments: visual-to-attribute, attribute-to-visual and visual-to-attribute. E-PGN \cite{yu2020episode} integrates the meta-learning approach into ZSL by formulating both the visual prototype generation and the class semantic inference into an adversarial framework. TF-VAEGAN \cite{narayan2020latent} proposes a feedback module in a VAE-GAN model to modulate the latent representation of the generator. However, the CNN visual features contain semantic-unrelated information, \textit{e.g.,} background noise and unannotated characteristics, which may jeopardise the semantic-visual alignment, we propose to factorize out the semantic-unrelated features and leverage the remaining semantic-consistent features as the generation target. 


\begin{figure*}
    \centering
    \includegraphics[width=0.92 \textwidth]{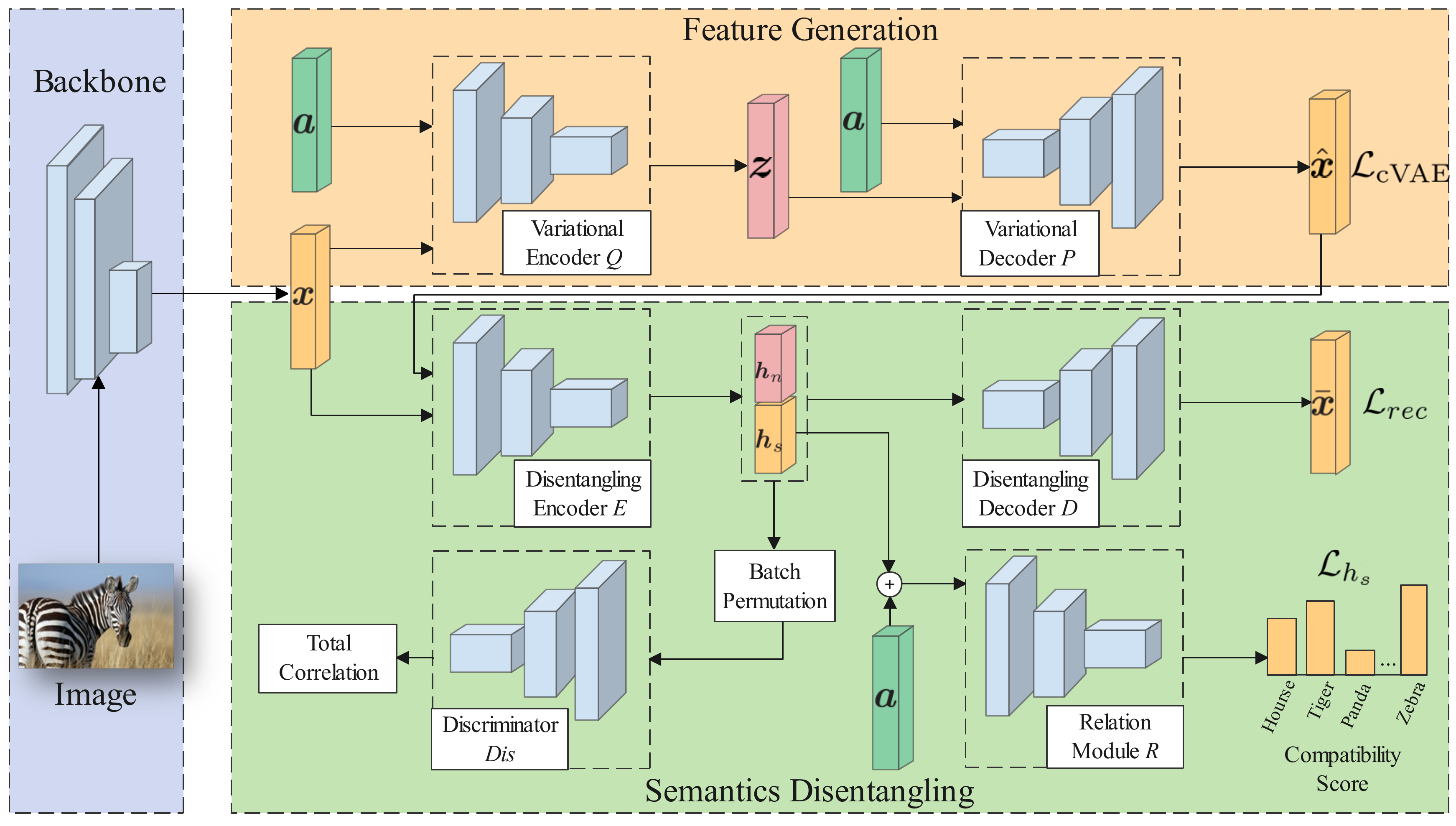}
    \caption{
    An illustration of our proposed SDGZSL, which is comprised of: (\romannumeral 1) a conditional VAE model for visual feature generation (in the orange box); (\romannumeral 2) disentangling modules to learn two factorized latent representations (in the green box). Particularly, the inference network $Q$ is learned to infer a latent variable $\vz$ from the visual feature $\vx$ and semantic embedding $\va$ of seen samples, where $\vz$ is then applied to recover unseen visual features. The encoder $E$ is subsequently trained to factorize the semantic-consistent $\vh_s$ and semantic-unrelated $\vh_n$ representations, and the independence between them is guaranteed by a total correlation penalty. By concatenating $\vh_s$ with the random semantic vectors $\va$, the derived relation module $R$ distinguishes the matched pairs from the mismatched ones, thus forcing $\vh_s$ to be semantically correlated.} 
    \label{model}
\end{figure*}
In most literature, disentanglement refers to independence among features of one representation \cite{bengio2013representation, eastwood2018framework, higgins2016beta, chen2018isolating}. 
Total correlation \cite{watanabe1960information} is a measurement of independence for multiple random variables. In information theory, total correlation is one of many generalizations of mutual information to random variables. It has been a key component in recent methods towards disentanglement. FactorVAE \cite{kim2018disentangling} is proposed to disentangle features by making the distribution of representations to be factorial and thus achieves independence across dimensions. Specifically, dimension-wise independence is achieved by applying the total correlation penalty on an original representation and its random permuted representation across dimensions. In contrast, we aim to enforce independence between two representations rather than keeping each dimension independent to each other. 
To our best knowledge, DLFZRL \cite{tong2019hierarchical} is the only work proposed to consider feature disentangling for ZSL. DLFZRL proposes a hierarchical disentangling approach to learn the discriminative latent features. This approach is designed in a two-step fashion, starting with feature selection and then learning to generate the selected features. The original visual features are factorized into three latent representations, including semantic, non-semantic, and non-discriminative features. 
However, we argue that the non-semantic visual features do not align with the semantic embeddings. Hence, it is hard to transfer the non-semantic visual features from seen to unseen classes. In our approach, we filter out such non-semantic visual features. Moreover, we combine the generative model and the disentangling modules so that our approach can be trained in an end-to-end manner.

\section{Proposed Approach}

This section firstly introduces the problem formulation and notations, then depicts the proposed approach of factorizing semantic-consistent representations for GZSL.

\subsection{Preliminaries}
Let $\{\mathcal{X}^s, \mathcal{Y}^s\}$ be the dataset with $S$ seen classes, which contains $N^s$ training samples $\mathcal{X}^s$ = $\{ \vx^s_{(i)}\}_{i=1}^{N^s}$ and the corresponding class labels $\mathcal{Y}^s$ = $\{ y^s_{(i)}\}_{i=1}^{N^s}$. The class labels span from 1 to $S$, $y^s \in \bm{L}^s = \{1,...,S\}$. Given another dataset $\{\mathcal{X}^u, \mathcal{Y}^u\}$, in which the classes are related to the seen dataset (\textit{e.g.,} all classes in both datasets correspond to animals). The dataset has $U$ unseen classes and consists of $N^u$ data instances $\mathcal{X}^u = \{\vx^u_{(j)}\}_{j=1}^{N^u}$ with the corresponding labels $\mathcal{Y}^u = \{y^u_{(j)}\}_{j=1}^{N^u}$. The class labels thus range from $S+1$ to $S+U$, $y^u \in \bm{L}^u = \{S+1, ..., S+U\}$.  The seen and unseen classes are mutually exclusive, \textit{i.e.,} $\bm{L}^s \cap \bm{L}^u = \varnothing$. For the semantic information $\mathcal{A} = \{\va_{(k)}\}^{S+U}_{k=1}$, each class from both seen and unseen is associated with a class-level semantic vector, which can be embeddings or attributes. We denote $\mathcal{A}^{s}$ and $\mathcal{A}^{u}$ as the semantic vectors of seen and unseen classes. 

\subsection{Semantics Disentanglement}
We start with defining three concepts in GZSL concerning the visual features and the semantic information.

\noindent\textit{\textbf{Semantic-consistent Representations}.} We define semantic-consistent representations $\vh_s$ to represent the characteristics of images annotated with attributes. The visual features of images are extracted by a deep model, \textit{e.g.,} ResNet101, pre-trained on ImageNet. These visual features are not task-specific for ZSL datasets, in which the classes are usually related (\textit{e.g.,} they all correspond to birds). Thus, the extracted visual features may unexpectedly involve redundant information that could compromise the semantic-visual relationship learning. The elegant solution to ZSL is attribute-based learning, which consists in introducing the intermediate semantic space $\mathcal{A}$. Such a space enables parameter sharing between classes. Ideally, if the visual features only contain information that corresponds to the annotated attributes $\vh_s$, the visual-semantic relation can be appropriately learned from seen classes and further transferred to unseen classes which could be beneficial to the GZSL task.

\noindent\textit{\textbf{Semantic-unrelated Representations}.} In analogy to $\vh_s$, we define semantic-unrelated visual representations $\vh_n$ to represent the information contained in visual features which may help classification but not correspond to the annotated attributes, \textit{e.g.,} in Figure \ref{att} the ear shape is an intuitively discriminative characteristic that can help distinguish animals but does not appear in the annotated attributes.

\noindent\textit{\textbf{Independence between $\vh_s$ and $\vh_n$}.}
In GZSL, the core mission is to transfer the semantic-visual relationship from seen classes to unseen classes. However, as the visual concepts of $\vh_n$ are not consistent with the manually annotated semantic information, it is difficult to align the relationship from semantic space to visual space and further transfer to unseen classes. As in the definitions of $\vh_s$ and $\vh_n$, in the setting of GZSL, the visual features can be disentangled into semantic-consistent $\vh_s$ and semantic-unrelated $\vh_n$ representations. To achieve this, we enforce the independence between $\vh_s$ and $\vh_n$. Independence can be measured by the mutual information, and the total correlation is one of the generalizations of the mutual information. 
Thus, we aim to minimize the entanglement of the visual information in the two representations by leveraging the total correlation to measure how $\vh_s$ and $\vh_n$ are independent.


\subsubsection{Visual Feature Preservation by Reconstruction}
The disentangling modules are developed with an encoder-decoder architecture, coupled with a total correlation penalty and a relation module. 
An encoder $E_{\psi}$ parameterized with $\psi$ is adopted to map a visual feature vector $\vx$ to a latent representation $\vh$, \textit{i.e.,} $E_{\psi}$: $\mathbb{R}^{d} \rightarrow \mathbb{R}^{l+m}$, where $l$ and $m$ are the dimensions of the semantic-consistent and semantic-unrelated representations.
Then, we have:
\begin{equation}
\label{autoencoder}
          E_{\psi}(\vx) = \vh = [\vh_s, \vh_n].
\end{equation}
In order to preserve the visual features in $\vh$, a decoder $D_{\omega}$ is learned to transform $\vh$ to the original visual feature $\vx$, \textit{i.e.,} $D_{\omega}$: $\mathbb{R}^{l+m} \rightarrow \mathbb{R}^d$.
The reconstruction objective can be formulated as: 
\begin{flalign}
\label{rec}
    \mathcal{L}_{rec} = \sum_{\vx \in \mathcal{X}^{s}} \lVert \vx - D_{\omega}(\vh_{s},\vh_{n}) \rVert^2,
\end{flalign}
where we calculate the mean squared error between the original visual features $\vx$ and the reconstructed visual features $\bar\vx = D_{\omega}(\vh_{s},\vh_{n})$.

\subsubsection{Semantic-Visual Relationship Learning}
A relation network (RN) \cite{sung2018learning} is adopted to learn the semantic-consistent representations $\vh_s$ by maximizing the compatibility score (CS) between a latent representation $\vh_{s}$ and the corresponding semantic embeddings $\va^{s}$.
The relation module $R_{\kappa}$ learns the pair-wise relationship between the latent representations and semantic vectors. The input of $R_{\kappa}$ is the pairs of a latent representation $\vh_s$ and $N^c$ unique semantic embeddings $\mathcal{A}^{batch} =  \{\va_{(c)}\}_{c=1}^{N^c}$ from a training batch with $B$ training instances. The ground-truth CS of the matched pairs are set to be 1; the mismatched pairs are 0, which can be formulated as:
\begin{equation}
\label{cs}
CS(\vh_{(t)},\va_{(c)}) = 
\begin{cases}
0, \quad y_{(t)} \neq y_{(c)} 
\\
1, \quad y_{(t)} = y_{(c)}
\end{cases}
,
\end{equation}
where $t$ and $c$ refer to the $t$-th semantic-consistent representation and $c$-th unique semantic embedding in a training batch, $y_{(t)}$ and $y_{(c)}$ denote the class label of $\vh_{(t)}$ and $\va_{(c)}$.

With the CS defined in Equation \ref{cs}, we now formulate the loss function for the semantics embeddings $\mathcal{A}^{batch}$ and semantic-consistent information.
The relation module $R_{\kappa}$ with a Sigmoid activation function outputs the learned compatibility score between 0 and 1 for each pair.
The loss function for optimizing $\vh_s$ can then be formulated as:
\begin{flalign}
\label{hs}
\mathcal{L}_{\vh_s} &= \sum_{t=1}^{B} \sum_{c=1}^{N^c} \lVert R_{\kappa}(\vh_{s(t)},\va_{{(c)}}) - CS(\vh_{s(t)},\va_{{(c)}})\rVert^2,
\end{flalign}
where we calculate the mean squared error between the output compatibility score of each pair of $\vh_{s(t)}$ and $\va_{(c)}$ and the ground-truth in every single training batch. Normally, we have $N^c \leq B$ as one class that may contain many sampled visual feature vectors in one batch. 


\subsubsection{Disentanglement by Total Correlation Penalty}
To encourage the disentanglement between semantic-consistent $\vh_s$ and semantic-unrelated $\vh_n$ representations, we introduce a \textit{total correlation} penalty in our proposed method.
Within the encoding procedure, the latent representation $\vh\sim \gamma(\vh \mid \vx)$ is expected to contain both the semantic-consistent and the semantic-unrelated information. Therefore, the disentanglement between these two streams of information is crucial for meaningful representation learning. Along with the relation network that facilitates the learning of semantic-consistent representation $\vh_s$, we aim to make $\vh_n$ semantic-unrelated by encouraging the disentanglement between $\vh_n$ and $\vh_s$. From a probabilistic point of view, we can think both of them come from different conditional distributions:
\begin{equation}
        \vh_s \sim \gamma_1( \vh_s \mid \vx),\quad \vh_n \sim \gamma_2( \vh_n \mid \vx).
\end{equation}
Hence, for the two latent representations, the total correlation can be formulated as:
\begin{equation}
      TC=\text{KL}\left(\gamma||\gamma_1\cdot \gamma_2\right),
\end{equation}
where $\gamma$ $\vcentcolon=$ $\gamma( \vh_s, \vh_n\mid \vx)$ is the joint conditional probability of $\vh_s$ and $\vh_n$. In order to efficiently approximate the total correlation, we apply the density ratio estimation to distinguish the samples from the two distributions in an adversarial manner. A discriminator $Dis_{\varphi}$ is constructed to output an estimate of the probability $Dis_{\varphi}(\vh)$ whose input is independent. Thus, 
\begin{flalign}
\label{eq:sensity}
        TC = \mathbb{E}_\gamma \left (\log \frac{\gamma}{\gamma_1 \cdot \gamma_2} \right )  
         \approx \mathbb{E}_\gamma \left( \log \frac{Dis(\vh)}{1-Dis(\vh)} \right ),
\end{flalign}

where the approximation is derived in Appendix \ref{density}. Simultaneously, we train the discriminator $Dis_\varphi$ to maximize the probability of assigning the correct label to $\vh$ and $\tilde{\vh}$:
\begin{flalign}
\label{dis}
\mathcal{L}_{dis} = \log Dis_\varphi(\vh) + \log (1-Dis_\varphi(\tilde{\vh})),
\end{flalign}
where $\tilde{\vh}$ is the result by randomly permuting each $\vh_s$ and $\vh_n$ across the batch. The permutation process is described as following:
(1) Given a batch of latent representations $\{\vh_{(t)}\}_{t=1}^{B}$; (2) they are split into $\{\vh_{s(t)}\}_{t=1}^{B}$ and $\{\vh_{n(t)}\}_{t=1}^{B}$; (3) for both $\vh_s$ and $\vh_n$, we permute the batch indices twice on $\mathcal{B} = \{1,...,B\}$, yielding $\mathcal{B}'$ and $\mathcal{B}''$; (4)  with the permuted indices, reorder the latent representations $\{\vh_{s\mathcal{B}'(t)}\}_{t=1}^{B}$, $\{\vh_{n\mathcal{B}''(t)}\}_{t=1}^{B}$ and concatenate them into $\{\tilde{\vh}_{(t)}\}_{t=1}^{B}$.

\subsection{Visual Feature Generation}
To model the distribution of visual features conditioned on the semantic information, we leverage the conditional variational auto-encoder (cVAE) \cite{sohn2015learning} as the generative model. 
In GZSL, we aim to transfer the knowledge from seen classes to some other unseen classes. Hence, we represent the category information in cVAE as the class embeddings $\va$ to enable parameter sharing between classes. The objective function of the cVAE in our framework can then be written as:
\begin{flalign}
\label{vae}
   \mathcal{L}_{\text{cVAE}} = - ~ \textit{KL} ~ [q_{\phi} (\vz|\vx,\va) || p_{\theta} (\vz|\va)] \nonumber \\ + ~ \mathbb{E}_{q_{\phi} (\vz|\vx,\va)} [\log p_{\theta} (\vx|\vz, \va)],
\end{flalign}
where the first term is the KL divergence between two distributions $q_{\phi}(\vz|\vx,\va)$ and $p_{\theta}(\vz|\va)$ and the second term is the reconstruction loss. As illustrated in Figure \ref{model}, we use $Q_{\phi}$ and $P_{\theta}$ to denote the inference and generator networks to model $q_{\phi}(\vz|\vx,\va)$ and $p_{\theta}(\vz|\va)$, respectively. Specifically, given the visual features $\vx$ and the semantic embeddings $\va$, the inference network $Q_{\phi}$ produces the latent variables $\vz$. 
The generator network $P_{\theta}$ leverages the inferred latent variable $\vz$ and the class embeddings $\va$ to reconstruct the visual features $\hat{\vx}$. The reconstructed and original visual features $\hat{\vx}$, $\vx$ are fed into the disentangling module.

{\begin{algorithm}[t]
\hspace{-0.01em}\textbf{Input:} $\{\mathcal{X}^s, \mathcal{Y}^s\}$, $\mathcal{A}^{s}$, learning rate $\lambda$ \\
\hspace{-0.01em}\textbf{Initialize:} $\varphi$ and $W = \{\phi, \theta, \psi, \omega, \kappa\}$
\caption{SDGZSL training}\label{euclid}
\hrule
\begin{algorithmic}[1]
\While{not converged}
    \State Randomly select a batch $\{\vx_{(t)}^s, \vy_{(t)}^s\}_{t=1}^{B}$, $\{\va_{(c)}^{s}\}_{c=1}^{N^c}$
    \For{step = 0,...,$n_{dis}$}
        \State Compute $\bar{\vh}$ and $\vh$ with $\phi, \theta, \psi$ by Eq. {\color{red} 1}  
        \State Vector permutation for $\bar{\vh}$ and $\vh$
        \State Compute $\lambda_3 \mathcal{L}_{dis}$ by Eq. {\color{red} 8} 
        \State Update $\varphi \leftarrow \varphi+\lambda\nabla_{\varphi} \lambda_3 \mathcal{L}_{dis}$
        \State Compute $\mathcal{L}_{overall1}$ = $\mathcal{L}_{\text{cVAE}}$ + $\mathcal{L}_{rec}$ + $\lambda_1 \mathcal{L}_{\vh_{s}}$  with Eq. {\color{red} 9},  {\color{red} 2}, {\color{red} 4} 
        \State Update $W \leftarrow W +\lambda\nabla_{W}\mathcal{L}_{\text{overall1}}$
    \EndFor

    \State Randomly select a batch $\{\vx_{(t)}^s, \vy_{(t)}^s\}_{t=1}^{B}$, $\{\va_{(c)}^{s}\}_{c=1}^{N^c}$
    \State Compute $\mathcal{L}_{overall2}$ = $\mathcal{L}_{\text{cVAE}}$ + $\mathcal{L}_{rec}$ + $\lambda_1 \mathcal{L}_{\vh_{s}}$ + $\lambda_2 TC$ with Eq. {\color{red} 9}, {\color{red} 2}, {\color{red} 4}, {\color{red} 7}  
    \State Update $W \leftarrow W+\lambda\nabla_{W}\mathcal{L}_{\text{overall2}}$
\EndWhile
\vspace{-1em}
\end{algorithmic}
\hrulefill \\
\hspace{-0.01em} \textbf{Output:} trained generative network $P_{\theta}$ and encoder $E_{\psi}$
\end{algorithm}\vspace{-5pt}}

\vspace{-5pt}
\subsection{Training and Inference}
\label{training}
Algorithm 1 shows the pseudocode of the model training. We iteratively train the $Dis_{\varphi}$ with the overall framework for $n_{dis}$ steps and then fix the weights in $Dis_{\varphi}$ to train other components. In Algorithm 1, the weights for $\mathcal{L}_{\vh_s}, TC$, and $\mathcal{L}_{dis}$ are denoted as $\lambda_1, \lambda_2$ and $\lambda_3$.
Once the training of SDGZSL is converged, the semantic-consistent representations $\bar{\vh}_s^{u}$ of unseen classes can be generated by the generative network $P_{\theta}$ and disentangling encoder $E_{\psi}$ from the Gaussian noise $\vz$ and the unseen semantic embeddings $\va^{u}$, \textit{i.e.,} $P_{\theta}: \mathbb{R}^{z} \times \mathbb{R}^{k} \rightarrow \mathbb{R}^{d}$, $E_{\psi}: \mathbb{R}^{d} \rightarrow \mathbb{R}^{l+m}$. $z$ represents the dimensions of the latent variables $\vz$. Then, we disentangle the training seen features $\{\vx_{(i)}^{s}\}_{i=1}^{N^s}$ into $\{\vh_{s(i)}^{s}\}_{i=1}^{N^s}$, together with the generated unseen semantic-consistent representations $\bar{\vh}_s^{u}$ we can simply train a supervised classifier. Further prediction for either seen or unseen objects can be conducted by supervised classification. In our paper, a Softmax classifier is adopted for evaluation.

\begin {table*}[t]
\caption {Performance comparison in accuracy (\%) on four datasets. We report the accuracies of unseen, seen classes and their harmonic mean for GZSL, which are denoted as U, S and H. For ZSL, performance results are reported with the average top-1 classification accuracy (T1). The top two results of the T1 and H are highlighted in bold. $\dag$ and $\ddag$ represent embedding-based and generative methods, respectively. $^*$ means a fine-tuned backbone is used. }
\centering
\scalebox{0.8}{
\begin{tabular}[t]{c|c|cccc|cccc|cccc|cccc}
\specialrule{.1em}{.00em}{.00em}

 \rowcolor[gray]{.9} \multirow{2}{*}{} &  \multirow{2}{*}{}   & \multicolumn{4}{c|}{aPaY} & \multicolumn{4}{c|}{AWA}  &  \multicolumn{4}{c|}{CUB} & \multicolumn{4}{c}{FLO}  \\  \hhline{>{\arrayrulecolor [gray]{0.9}}-->{\arrayrulecolor {black}}----------------} 
  
 \rowcolor[gray]{.9} &  \multirow{-2}{*}{Methods} & \multicolumn{1}{c|}{\textit{T1}} & \textit{U} & \textit{S} & \textit{H}& \multicolumn{1}{c|}{\textit{T1}}  & \textit{U} & \textit{S} & \textit{H} & \multicolumn{1}{c|}{\textit{T1}} & \textit{U} & \textit{S} & \textit{H} & \multicolumn{1}{c|}{\textit{T1}} & \textit{U} & \textit{S} & \textit{H}\\

  \specialrule{.1em}{.00em}{.00em}
\multirow{6}{*}{\dag}   
                    & LFGAA       \cite{liu2019attribute}   
&\multicolumn{1}{c|}{-}    & -             & -                 & -       
&\multicolumn{1}{c|}{68.1}    & 27.0          & 93.4              & 41.9  
&\multicolumn{1}{c|}{67.6}    & 36.2          & 80.9              & 50.0      
&\multicolumn{1}{c|}{-}    & -             & -                 & -    
    \\

                    & DCN       \cite{liu2018generalized}
&\multicolumn{1}{c|}{43.6}    & 14.2          & 75.0              & 23.9       
&\multicolumn{1}{c|}{65.2}    & 25.5          & 84.2              & 39.1  
&\multicolumn{1}{c|}{56.2}    & 28.4          & 60.7              & 38.7      
&\multicolumn{1}{c|}{-}        & -             & -                 & -    
    \\
                    & TCN       \cite{jiang2019transferable}
&\multicolumn{1}{c|}{38.9}    & 24.1          & 64.0              & 35.1       
&\multicolumn{1}{c|}{71.2}    & 61.2          & 65.8              & 63.4
&\multicolumn{1}{c|}{59.5}    & 52.6          & 52.0              & 52.3      
&\multicolumn{1}{c|}{-}    & -             & -                 & -    
    \\
                    & DVBE      \cite{min2020domain}
&\multicolumn{1}{c|}{-}    & 32.6          & 58.3              & 41.8          
&\multicolumn{1}{c|}{-}    & 63.6          & 70.8              & 67.0     
&\multicolumn{1}{c|}{-}    & 53.2          & 60.2              & 56.5         
&\multicolumn{1}{c|}{-}    & -             & -                 & -   
    \\

 \specialrule{.1em}{.20em}{.20em}
\multirow{10}{*}{\ddag} 

                        & f-CLSWGAN \cite{xian2018feature}     
&\multicolumn{1}{c|}{40.5}   & 32.9           & 61.7              & 42.9    
&\multicolumn{1}{c|}{65.3}   & 56.1           & 65.5              & 60.4     
&\multicolumn{1}{c|}{57.3}   & 43.7           & 57.7              & 49.7	        
&\multicolumn{1}{c|}{69.6}   & 59.0           & 73.8              & 65.6 
   \\
 &  CANZSL   \cite{chen2020canzsl}         
&\multicolumn{1}{c|}{-}    & -                 & -                 & -  
&\multicolumn{1}{c|}{68.9}    & 49.7              & 70.2              & 58.2  
&\multicolumn{1}{c|}{60.6}    & 47.9              & 58.1              & 52.5 
&\multicolumn{1}{c|}{69.7}    & 58.2              & 77.6              & 66.5 
    \\
  &  LisGAN   \cite{li2019leveraging}         
&\multicolumn{1}{c|}{43.1}    & 34.3        & 68.2               & 45.7  
&\multicolumn{1}{c|}{70.6}    & 52.6              &76.3              & 62.3  
&\multicolumn{1}{c|}{58.8}    & 46.5              & 57.9              & 51.6 
&\multicolumn{1}{c|}{69.6}    & 57.7              & 83.8              & 68.3 
    \\
                    & CADA-VAE  \cite{schonfeld2019generalized}
&\multicolumn{1}{c|}{-}    & 31.7              & 55.1              & 40.3          
&\multicolumn{1}{c|}{64.0}    & 55.8              & 75.0              & 63.9     
&\multicolumn{1}{c|}{60.4}    & 51.6              & 53.5              & 52.4     
&\multicolumn{1}{c|}{65.2}    & 51.6              & 75.6              & 61.3 
    \\        
                    & f-VAEGAN-D2  \cite{xian2019f} 
&\multicolumn{1}{c|}{-}    & -                 & -                 & -         
&\multicolumn{1}{c|}{71.1}    & 57.6              & 70.6              & 63.5     
&\multicolumn{1}{c|}{61.0}    & 48.4              & 60.1              & 53.6     
&\multicolumn{1}{c|}{67.7}    &56.8              & 74.9              & 64.6 
    \\       
                     & DLFZRL  \cite{tong2019hierarchical}
&\multicolumn{1}{c|}{\textbf{46.7}}    & -                 & -                 & 38.5          
&\multicolumn{1}{c|}{70.3}    & -                 & -                 & 60.9     
&\multicolumn{1}{c|}{61.8}    & -                 & -                 & 51.9     
&\multicolumn{1}{c|}{-}    & -                 & -                 & -   
    \\       
             
                    & TF-VAEGAN  \cite{narayan2020latent}
&\multicolumn{1}{c|}{-}    & -                 & -                 & -     
&\multicolumn{1}{c|}{72.2}    & 59.8              & 75.1              &  66.6  
&\multicolumn{1}{c|}{64.9}    & 52.8              & 64.7              & 58.1       
&\multicolumn{1}{c|}{70.8}    & 62.5              & 84.1              & 71.7  
    \\
     & TF-VAEGAN$^{*}$  \cite{narayan2020latent}
&\multicolumn{1}{c|}{-}    & -                 & -                 & -     
&\multicolumn{1}{c|}{73.4}    & 55.5              & 83.6              & 66.7  
&\multicolumn{1}{c|}{74.3}    & 63.8              & 79.3              & 70.7       
&\multicolumn{1}{c|}{74.7}    & 69.5              & 92.5              & 79.4  
    \\
                    & OCD-CVAE  \cite{keshari2020generalized}
&\multicolumn{1}{c|}{-}    & -                 & -                 & -     
&\multicolumn{1}{c|}{71.3}    & 59.5              & 73.4              &  65.7 
&\multicolumn{1}{c|}{60.3}    & 44.8              & 59.9              & 51.3       
&\multicolumn{1}{c|}{-}    & -                 & -                 & -   
    \\
                    & E-PGN  \cite{yu2020episode}
&\multicolumn{1}{c|}{-}    & -                 & -                 & -          
&\multicolumn{1}{c|}{73.4}    & 52.6              & 83.5              & 64.6    
&\multicolumn{1}{c|}{72.4}    & 52.0              & 61.1              & 56.2         
&\multicolumn{1}{c|}{\textbf{85.7}}    & 71.5              & 82.2              & 76.5  
     
    \\ 
     & AGZSL  \cite{yu2020episode}
&\multicolumn{1}{c|}{41.0}    & 35.1            & 65.5              & \textbf{45.7}          
&\multicolumn{1}{c|}{73.8}    & 65.1              & 78.9              & 71.3    
&\multicolumn{1}{c|}{57.2}    & 41.4              & 49.7              & 45.2         
&\multicolumn{1}{c|}{82.7}    & 63.5              & 94.0              & 75.7  
    \\
         & AGZSL$^{*}$  \cite{yu2020episode}
&\multicolumn{1}{c|}{43.7}    & 36.2        & 58.6                 & 44.8          
&\multicolumn{1}{c|}{\textbf{76.4}}    & 69.0              & 86.5              & \textbf{76.8}    
&\multicolumn{1}{c|}{\textbf{77.2}}    & 69.2              & 76.4              & \textbf{72.6}         
&\multicolumn{1}{c|}{85.2}    & 73.7              & 91.9              & 81.7  
    \\
\specialrule{.1em}{.00em}{.00em}
\rowcolor[gray]{.9}  &  cVAE      
&\multicolumn{1}{c|}{39.2}            & 30.2      & 55.3     & 39.0      
&\multicolumn{1}{c|}{65.4}            & 54.4      & 72.6      & 62.2   
&\multicolumn{1}{c|}{61.4}            & 47.0      & 59.9      & 52.7      
&\multicolumn{1}{c|}{68.7}            & 60.1      & 89.6      & 71.9  
            \\

\rowcolor[gray]{.9}  &  SDGZSL w/o RN\&TC   
&\multicolumn{1}{c|}{20.3}            & 14.6      & 37.0      & 21.0       
&\multicolumn{1}{c|}{57.5}            & 41.3      & 70.0      & 51.9   
&\multicolumn{1}{c|}{42.3}            & 25.1      & 54.7      & 34.5         
&\multicolumn{1}{c|}{67.5}            & 56.6      & 87.6      & 68.7   
            \\
            
\rowcolor[gray]{.9}  &  SDGZSL w/o TC     
&\multicolumn{1}{c|}{39.8}            & 33.8      & 49.1      & 40.0        
&\multicolumn{1}{c|}{54.0}            & 45.7      & 79.4      & 58.0 
&\multicolumn{1}{c|}{27.4}            & 23.6      & 46.2      & 31.2  
&\multicolumn{1}{c|}{57.5}            & 44.8      & 65.7      & 53.3     
            \\
            
\rowcolor[gray]{.9}  &  SDGZSL w/o RN
&\multicolumn{1}{c|}{30.9}            & 27.3      & 41.9      & 33.1        
&\multicolumn{1}{c|}{55.8}            & 48.5      & 64.6      & 55.5  
&\multicolumn{1}{c|}{30.4}            & 20.8      & 31.4      & 25.0   
&\multicolumn{1}{c|}{53.2}            & 40.6      & 67.6      & 50.7
            \\
            
\rowcolor[gray]{.9} &   SDGZSL      
&\multicolumn{1}{c|}{45.4}            & 38.0 & 57.4 & \textbf{45.7}  
&\multicolumn{1}{c|}{72.1}            & 64.6 & 73.6 & 68.8
&\multicolumn{1}{c|}{75.5}            & 59.9 & 66.4 & 63.0
&\multicolumn{1}{c|}{85.4}            & 83.3 & 90.2 & \textbf{86.6} 
\\
\rowcolor[gray]{.9} &   SDGZSL$^{*}$      
&\multicolumn{1}{c|}{\textbf{47.0}}            & 39.1 & 60.7 & \textbf{47.5}  
&\multicolumn{1}{c|}{\textbf{74.3}}            & 69.6 & 78.2 & \textbf{73.7}
&\multicolumn{1}{c|}{\textbf{78.5}}            & 73.0 & 77.5 & \textbf{75.1}
&\multicolumn{1}{c|}{\textbf{86.9}}            & 86.1 & 89.1 & \textbf{87.8} 
\\
\specialrule{.1em}{.00em}{.00em}
\end{tabular}}
\label{gzslperoformance}
\end {table*} 

\section{Experiments}

\subsection{Experimental Setting}
\noindent \textbf{Datasets} The proposed framework is evaluated on four widely used benchmark datasets of image classification, including two coarse-grained datasets (Attribute Pascal and Yahoo (\textbf{aPaY}) \cite{farhadi2009describing} and Animals with Attributes 2 (\textbf{AWA}) \cite{lampert2013attribute}) and two fine-grained datasets (Caltech-UCSD Birds-200-2011 (\textbf{CUB}) \cite{wah2011caltech} and Oxford  Flowers (\textbf{FLO}) \cite{nilsback2008automated}). 
aPaY contains 18,627 images from 42 classes and is annotated with 64 attributes. It combines datasets a-Pascal and a-Yahoo, which has 30 and 12 classes respectively. 
AWA is a relatively larger coarse-grained dataset with 30,475 images from 50 animal species, in which 40 are selected as seen classes and the rest is unseen. 
Each species in the dataset is annotated with 85 attributes. CUB consists of 11,788 images from fine-grained bird species with 150 seen and 50 unseen classes. 
FLO contains 102 flower categories with 82 seen and 20 unseen classes. The semantic embeddings of FLO and CUB are 1,024-dimensional character-based CNN-RNN features \cite{reed2016learning} extracted from the fine-grained visual descriptions (10 sentences per image).

\noindent \textbf{Evaluation Protocol}
The metric used to evaluate the GZSL task is harmonic mean, which calculates the joint accuracy of the seen and unseen classes. The formula used to calculate the harmonic mean $\mathit{{H}}$ can be written as: $\mathit{H} = (2 \times \mathit{U}  \times \mathit{S}) / (\mathit{U} + \mathit{S})$, where $\mathit{U}$ and $\mathit{S}$ denote the average per-class top-1 accuracy of unseen and seen classes, respectively. A high harmonic mean indicates the good performance of both seen and unseen classes. 

\begin{figure}[t]
    \centering
    \includegraphics[width=0.9\linewidth]{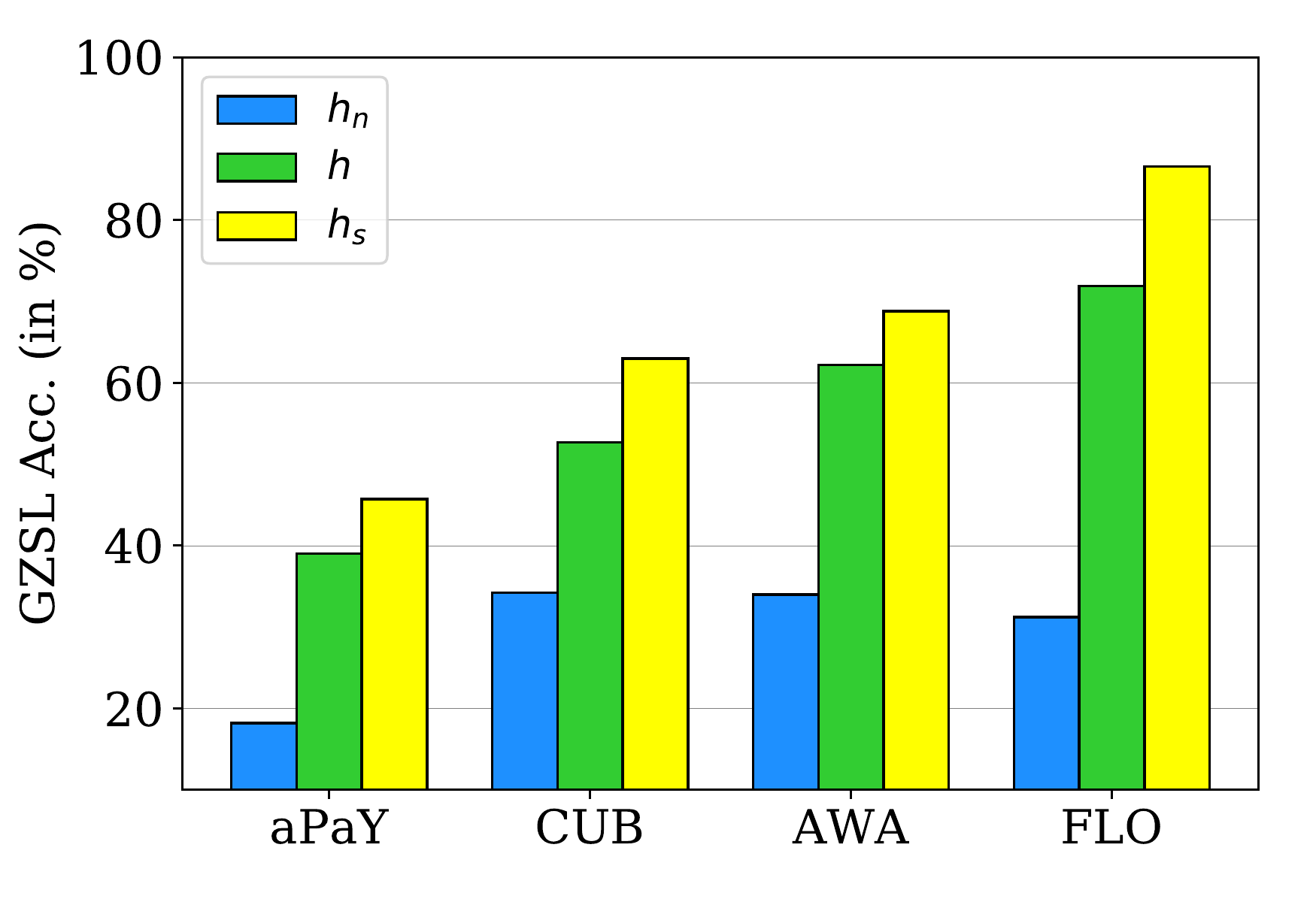}
    \caption{GZSL accuracy (in \%) comparison between $\vh_s$, $\vh_n$, and $\vh$.}
    \label{bar}
\end{figure}

\subsection{Comparison with State-of-the-art Methods}
Table \ref{gzslperoformance} shows the GZSL performance of compared methods and ours with and without fine-tuning the backbone on the datasets. We choose the recent state-of-the-art embedding-based and generative methods. They are marked with $\dag$ and $\ddag$ respectively. 
Generally, our proposed method consistently performs better than all the compared methods, except on AWA dataset. These methods directly use the visual features extracted from a pre-trained or fine-tuned ResNet101 model, which limits semantic-visual alignment learning. The intuition of this work is in agreement with DLFZRL \cite{tong2019hierarchical}, we both aim to disentangle more effective representations from visual features. 
DLFZRL proposes to disentangle the discriminative features from visual features, both semantic and non-semantic. However, we argue that the non-semantic discriminative features cannot be transferred from seen classes to unseen classes. The conditional generative models learn to project the semantic vectors to the visual space, but intuitively there is no way to generalize the non-semantic discriminative features from the semantic perspective. As can be seen from the performance comparison, our method surpasses DLFZRL on all the reported datasets. Our method is also applicable to the generative methods mentioned above. It is worth mentioning that, unlike DFLZRL that is designed in a two-stage process, we carefully design our framework by incorporating the disentangling modules into the generative model so that the overall framework can be trained in an end-to-end manner. 

\subsection{Conventional Zero-shot Learning Results}
We aim to solve the generalized zero-shot learning problem in this work, but it is necessary to demonstrate that our approach can also achieve state-of-the-art performance on conventional zero-shot learning that only aims to classify unseen class samples over unseen classes. The results shown in Table \ref{gzslperoformance} demonstrate the performance comparison between our proposed SDGZSL method and existing state-of-the-art models. It can be seen that our method performs better than all other methods on CUB and FLO datasets. Even if DLFZRL is slightly higher than our method on aPaY dataset, we can surpass it on all other datasets. This is the same case for E-PGN on dataset AWA. 

\begin{figure}[t]
    \centering
    \includegraphics[width=0.9\linewidth]{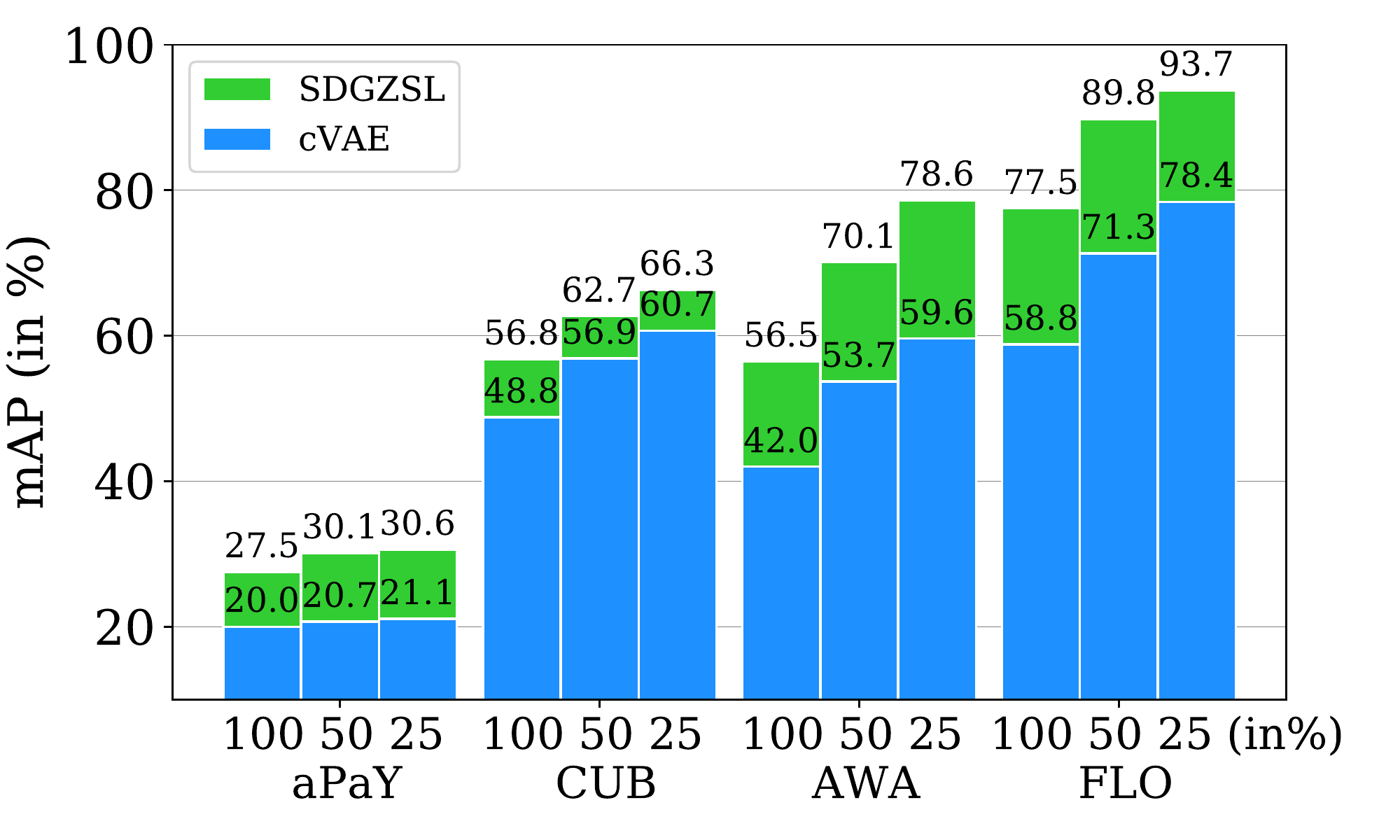}
    \caption{Zero-shot image retrieval result comparison between cVAE and SDGZSL.}
    \label{barretrieval}
\end{figure}

\begin{figure}[t]
    \centering
    \includegraphics[width=0.47\textwidth]{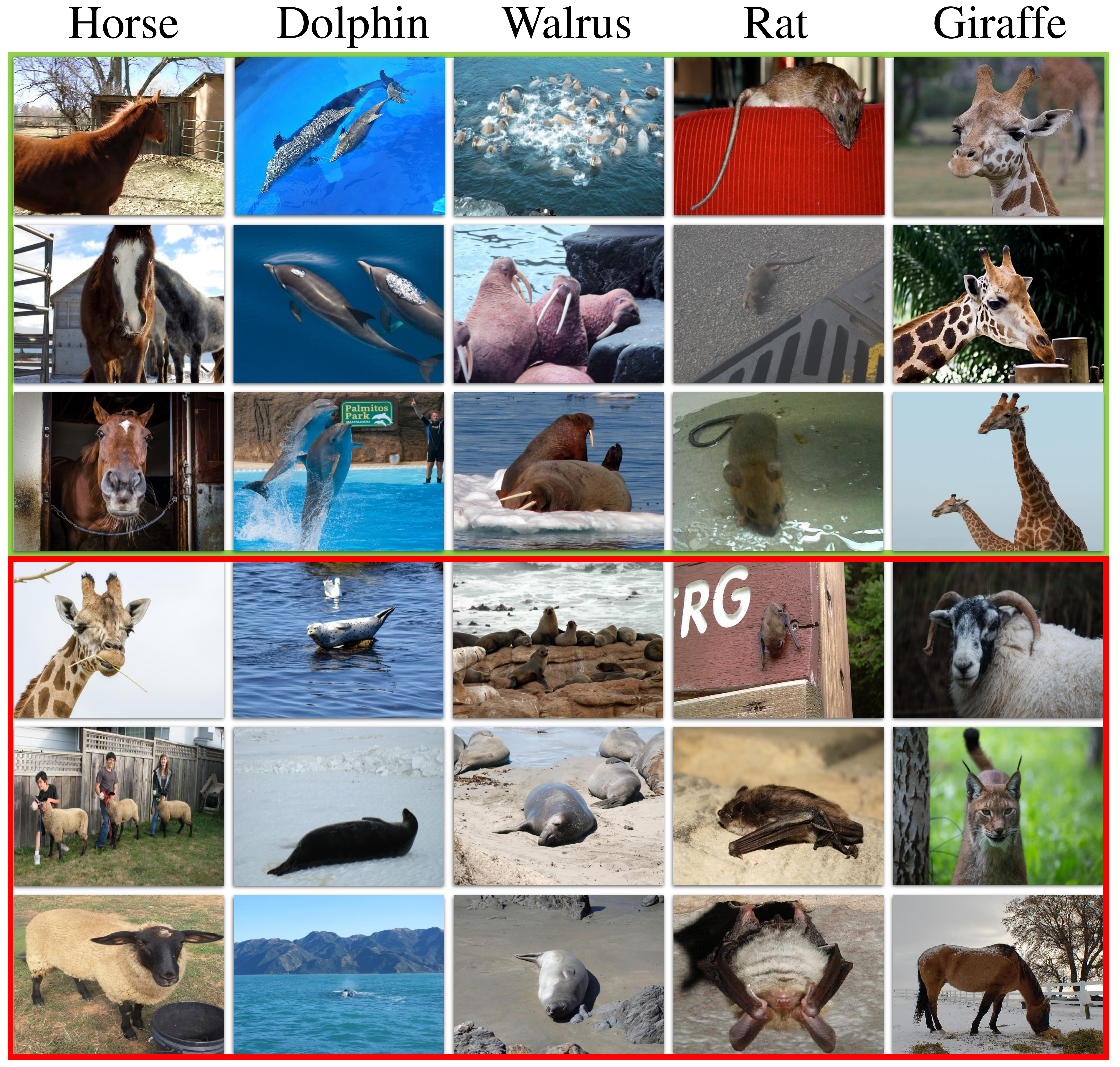}
    \caption{Qualitative results of our approach on AwA, where six random unseen class labels are shown on top. The top-3 retrieved images are highlighted in green and the top-3 retrieved false examples are highlighted in red.}
    \label{retrieval}
    \vspace{-12pt}
\end{figure}

\subsection{Zero-shot Retrieval Results}
We conduct the image retrieval task to illustrate the qualitative results of the proposed framework. Given the semantic embeddings of a specific class, we synthesize a certain number of semantic-consistent representations for a specific class and compute the centroid point as a retrieval query, which is further used to retrieve the nearest samples. To evaluate the performance on the retrieved samples, the mean average precision (mAP) score is adopted.
In Figure \ref{barretrieval}, we compare our proposed SDGZSL method with the base generative model cVAE when retrieving 100\%, 50\%, and 25\% of the images from all the unseen classes on aPaY, AWA, CUB, and FLO. It can be seen that the disentangling module can significantly boost the retrieval performance among all settings, which can also demonstrate the effectiveness of the disentangling modules from the retrieval perspective. Figure \ref{retrieval} illustrates the retrieved examples on the AWA dataset. The class name is given on top, following by the top-3 true positive retrieved images with green boxes and the top-3 false-positive retrieved examples with red boxes. It can be seen that all the false-positive images look very similar to the groundtruth examples. For example, the top-3 failed retrieved samples for the rat are all bats, as the two species have many visual patterns in common. Thus, the nearest neighbor-based retrieval may fail to distinguish the false positive cases. The results demonstrate that the synthesized semantic-consistent representations are close to the samples of the same class in the semantic-consistent feature space.

\begin{figure*}[t]
    \centering
    \includegraphics[width=1\textwidth]{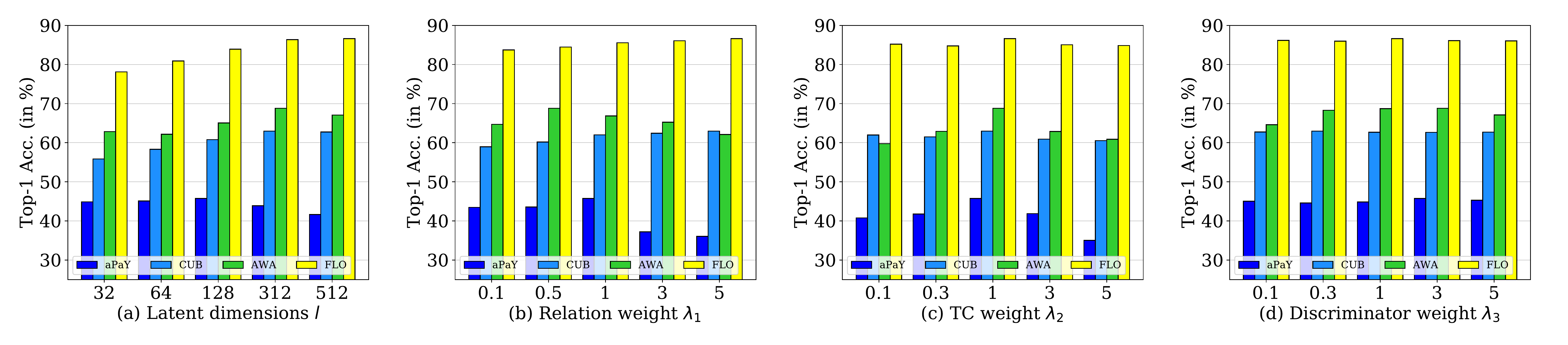}
    \caption{Hyper-parameter study w.r.t. latent dimensions $l$, relation weight $\lambda_1$, TC weight $\lambda_2$, and discriminator weight $\lambda_3$.}
    \label{hyper}
\end{figure*}

\subsection{Model Analysis}
\noindent \textbf{Ablation Study}.
In this ablation study, we evaluate various stripped-down versions of our full proposed model to validate the contributions of the key disentangling components of the model. In Table \ref{gzslperoformance}, we report the GZSL performance of each version on the four benchmark datasets. The best performance is achieved when RN and TC are both applied. 
In Figure \ref{bar}, we show the performance comparison between $\vh_s$, $\vh_n$, and $\vh$. 
From the observations on all datasets, semantic-unrelated features $\vh_n$ performs much worse than the original latent features $\vh$, which proofs the ineffectiveness of $\vh_n$ in GZSL. In contrast, using semantic-consistent features $\vh_s$ can further improve the performance, which validates the significant impact of $\vh_s$ when transferring the connection between semantic to semantic-unrelated representations from seen to unseen classes in GZSL. 
It is worth noting that when taking both $\vh_s$ and $\vh_n$ (\textit{i.e., $\vh$}) to classify test samples, the performance is similar to cVAE. 


\begin{figure}[t]
    \centering
    \includegraphics[width=0.45\textwidth]{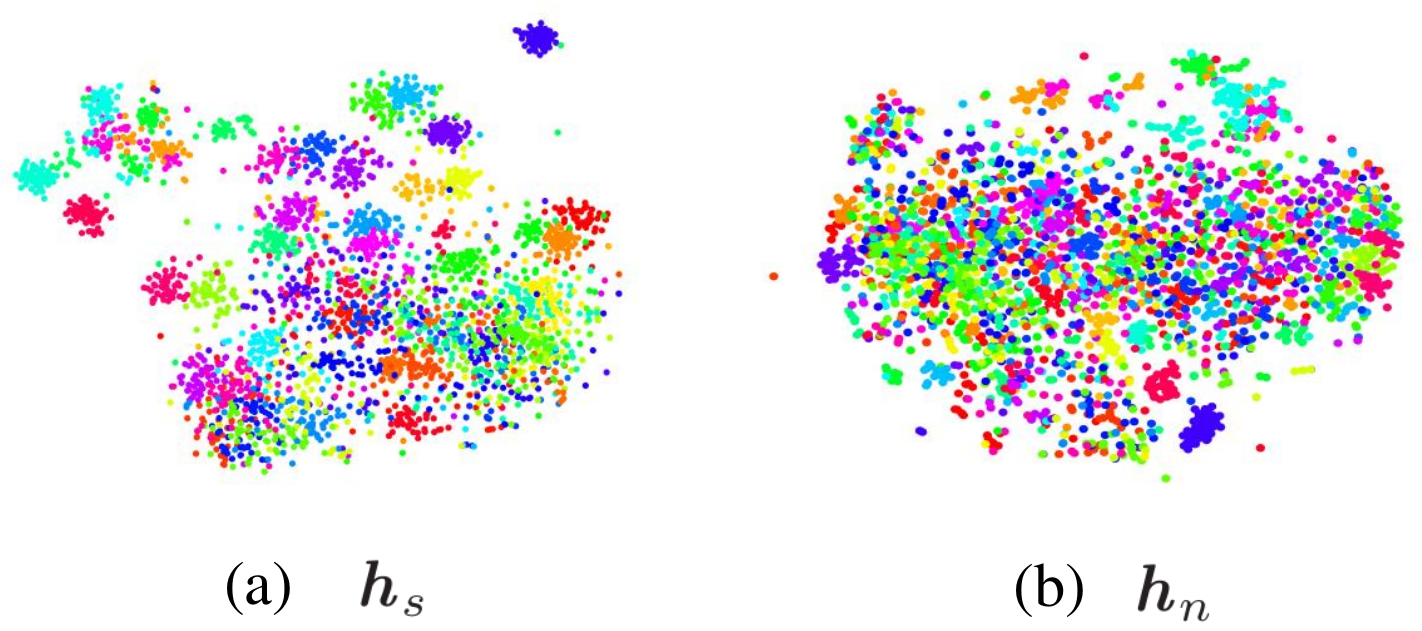}
    \caption{t-SNE visualization of different representations of 50 unseen classes on CUB: (a) semantic-consistent representations $\vh_s$; (b) semantic-unrelated representations $\vh_n$. }
    \label{tsne}
    \vspace{-10pt}
\end{figure}

\vspace{5pt}
\noindent \textbf{Hyper-Parameter Sensitivity}.
There are mainly four hyper-parameters that control the objective function, including the number of the disentangled feature dimension $\l$, the weight of \textit{total correlation} term $\lambda_1$, the weight of loss in the relation module $\lambda_2$, and the weight of the discriminator loss $\lambda_3$.
To better understand the effect of the disentangling components, we report the sensitivity of the four hyper-parameters in Figure \ref{hyper}.

\vspace{5pt}
\noindent \textbf{t-SNE Visualization}.
To further validate the properties of the disentanglement, we visualize semantic-consistent representations $\vh_s$ from $\vx$ in Figure \ref{tsne} (a) and semantic-unrelated representations $\vh_n$ from $\vx$ in (b). We choose all of 50 unseen classes from CUB dataset that have enough classes to show the class-wise comparison. Clearly, as we expected, the semantic-consistent representations $\vh_s$ is much more discriminative than the semantic-unrelated representations $\vh_n$. However, we can still see discriminative patterns from $\vh_n$ as there are remaining discriminative features even if they are semantic-unrelated, \textit{e.g.,} characteristics that are not annotated in the attributes. We argue that these discriminative features may help classify between these classes, but as the non-semantic discriminative features are not annotated in the attributes, it is intuitively impossible to transfer the semantic-visual relationship from seen classes to unseen classes.

\section{Conclusion}
In this paper, we propose a novel semantics disentangling approach for generalized zero-shot learning.
Specifically, the visual features of an image extracted from pre-trained ResNet101 are further factorized into two independent representations that are semantic-consistent and semantic-unrelated. In our approach, an encoder-decoder architecture is coupled with a relation module to learn the visual-semantic interaction. Further, we leverage the total correlation term to encourage the disentanglement between the two representations. The disentangling encoder-decoder model is incorporated into a conditional variational autoencoder and trained in an end-to-end manner. The generation ability trained on seen classes is transferred to the unseen classes and synthesizes the missing visual samples. We evaluate our proposed method on four image classification datasets. Extensive experiments show that our approach consistently performs better than other state-of-the-arts.

\vspace{5pt}
\noindent\textbf{Acknowledgments:}
This work was partially supported by Australian Research Council DP190101985, CE200100025, DE200101610 and Sichuan  Science and Technology Program under Grant 2020YFG0080.

{\small
\bibliographystyle{ieee_fullname}
\bibliography{files/bib/bibliography}
}

\newpage 
\appendix
\section*{Appendices}
\section{Total Correlation Approximation}
\label{density}
The density ratio of the total correlation term is:
\begin{equation}
    r(\vh)=\frac{\gamma}{\gamma_1 \cdot \gamma_2}.
\end{equation}

Let $\gamma_3$ denote $\gamma_1\cdot \gamma_2$. Assume there are equal numbers of pairs of $\vh_s$ and $\vh_n$, when $y=1$, $\vh_s$ and $\vh_n$ inter-dependent, and when $y=0$, $\vh_s$ and $\vh_n$ are independent. We have

\begin{flalign}
    r(\vh)&=\frac{\gamma}{\gamma_1 \cdot \gamma_2}\\
    &= \frac{\gamma}{\gamma_3}\\
    &= \frac{\tau (\vh \mid y=1)}{\tau(\vh\mid y=0)} \\
    &= \frac{\tau (y=1 \mid \vh) \tau (\vh) / \tau(y=1)} {\tau(y=0 \mid \vh) \tau (\vh) / \tau(y=0)}\\
    &= \frac{\tau(y=1\mid \vh)}{\tau(y=0\mid \vh)}\\
    &= \frac{\tau(y=1\mid \vh)}{1-\tau(y=1\mid \vh)}.
\end{flalign}

If we use a classifier/discriminator $Dis_\varphi(\vh)$ to approximate the term $\tau(y=1\mid \vh)$, we can write the density ratio above as:

\begin{equation}
    r(\vh)\approx\frac{Dis_\varphi(\vh)}{1-Dis_\varphi(\vh)}.
\end{equation}
Thus, the total correlation can be approximated by:
\begin{flalign}
\label{eq:sensity}
        TC \approx \mathbb{E}_\gamma \left( \log \frac{Dis_\varphi (\vh)}{1-Dis_\varphi(\vh)} \right ),
\end{flalign}

\begin{figure*}%
    \centering
    \subfloat[\centering cVAE]{{\includegraphics[width=\linewidth]{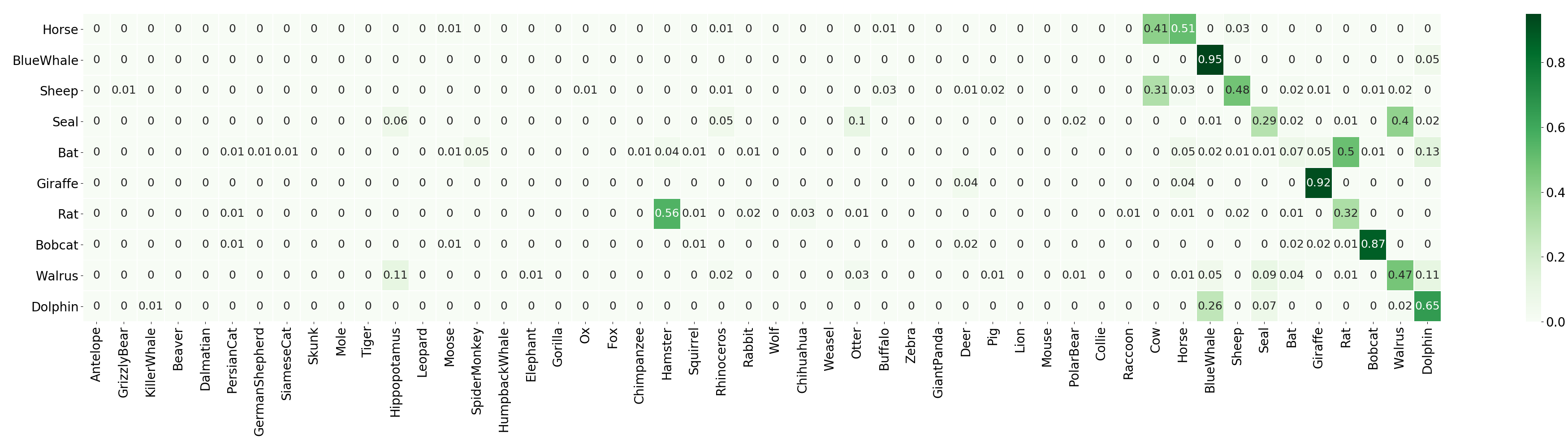} }}%
    \qquad
    \subfloat[\centering SDGZSL]{{\includegraphics[width=\linewidth]{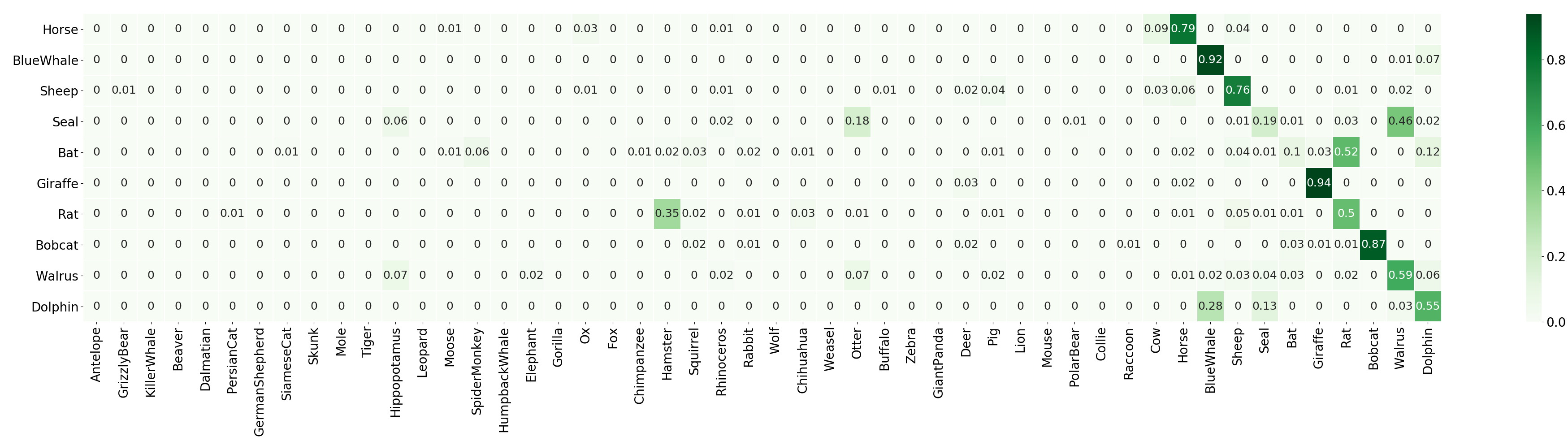} }}%
    \caption{The comparison between cVAE and the proposed method on confusion matrices of the unseen images being predicted over all classes. The vertical axis denotes the ground truth and the horizontal axis represents the predictions.   }%
    \label{cm}
\end{figure*}

\section{Implementation}
\label{implementation}
Our framework is implemented by the popular deep learning framework PyTorch 1.4. The disentangling encoder $E_{\psi}$ and decoder $D_{\omega}$ contain one and two fully connected (FC) layers, respectively. Each layer is followed a LeakyReLU activation function layer and a dropout layer. The numbers of hidden units of $E_{\psi}$ and $D_{\omega}$ are $[l+m]$ and $[2048, 2048]$, where  $l$ and $m$ are the numbers of dimensions of $\vh_s$ and $\vh_n$.  The relation module $R_{\kappa}$ is built with two FC layers, where the first layer is followed by a ReLU activation function layer and the second layer is followed by a Sigmoid activation function layer. The discriminator is implemented with a single FC layer followed by a Sigmoid activation function layer. We set the dimension number $l$ and $m$ as the same value between 32 and 512. The hyper-parameters for relation weight $\lambda_1$, TC weight $\lambda_2$, and discriminator loss $\lambda_3$ are set between 0.1 and 5. We use Adam optimizer with $\beta_{1}=0.9$, $\beta_{2}=0.999$ and batch size with 64. For the generative model cVAE, we use a five-layer MLP for encoder $Q_{\phi}$ and its structure can be written as: FC-LeakyReLU-FC-Dropout-LeakyReLU-FC for the first part; a single FC layer for the mean vector output; FC-Dropout-Softplus for the variance vector output. Another three-layer MLP for decoder $P_{\theta}$ can be written as: FC-ReLU-Dropout-FC-LeakyReLU. We warm up the KL term and the TC term gradually with increasing epochs.  All the experiments are performed on a Lenovo workstation with two NVIDIA GeForce GTX 2080 Ti GPUs.

\section{Class-wise Analysis}
To validate the merit of specific classes in our disentanglement approach, we compare the class-wise performance of unseen classes in AWA between the base generative method, \textit{i.e.,} a standard cVAE, and the proposed SDGZSL. As shown in Figure \ref{cm}, the top sub-figure is the confusion matrix of the cVAE while the bottom one is of the proposed SDGZSL. The rows represent the groundtruth labels of the target test samples while the columns represent the predicted labels of the test samples. As unseen classes are hard to achieve high performance and can be usually misclassified into seen classes, we take the test samples from unseen classes to compare between the two settings. There are 10 unseen classes in AWA dataset, and almost all these classes in SDGZSL gain higher accuracy than cVAE. The test samples of unseen classes are usually misclassified into visually similar classes. Notably, in cVAE we can see that 41\% of test samples from category "horse" are misclassified into "sheep", 31\% from "sheep" to "cow", 35\% from "rat" to "hamster". In contrast, our approach shows the ability to alleviate the problem by reducing the misclassification rate to 9\%, 3\%, and 35\% for categories "horse", "sheep" and "rat", respectively. However, some extremely hard categories, \textit{e.g.,} "seal" and "bat" can be easily misclassified into "walrus" and "rat", also fail in our proposed method. This will be investigated in our future work.

\begin {table*}[ht!]
\caption { Performance comparison in accuracy (\%) with traditional methods on four datasets. We report the accuracies of unseen, seen classes and their harmonic mean, which are denoted as $U$, $S$ and $H$. The best results of the harmonic mean are highlighted in bold.}
\begin{center}
\scalebox{0.7}{
\begin{tabular}[t]{c|ccc|ccc|ccc|ccc}

\specialrule{.1em}{.00em}{.10em}
  \rowcolor[gray]{.9}    & \multicolumn{3}{c|}{aPaY} & \multicolumn{3}{c|}{AWA}  &  \multicolumn{3}{c|}{CUB} & \multicolumn{3}{c}{FLO}  \\  
 \rowcolor[gray]{.9} Methods & \textit{U} & \textit{S} & \textit{H}  & \textit{U} & \textit{S} & \textit{H}  & \textit{U} & \textit{S} & \textit{H} & \textit{U} & \textit{S} & \textit{H} \\ 

  \specialrule{.1em}{.00em}{.00em}
                        DAP \cite{lampert2013attribute}
    & 4.8           & 78.3              & 9.0        
    & 0.9           & 84.7              & 0.0
    & 1.7           & 67.9              & 3.3   
    & -             & -                 & -     
    \\

                         LATEM   \cite{xian2016latent}  
    & 0.1           & 73.0              & 0.2        
    & 13.3          & 77.3              & 20.0
    & 15.2          & 57.3              & 24.0    
    & 6.6           & 47.6              & 11.5  
    \\
 
                         ALE    \cite{akata2015label}   
    & 4.6           & 73.7              & 8.7      
    & 14.0          & 81.8              & 23.9    
    & 23.7          & 62.8              & 34.4    
    & 13.3          & 61.6              & 21.9  
    \\
                        DeVise    \cite{frome2013devise}   
    & 3.5           & 78.4              & 6.7      
    & 17.1          & 74.7              & 27.8    
    & 23.8          & 53.0              & 32.8    
    & 13.2          & 82.6              & 22.8    
    \\
                      SJE       \cite{akata2015evaluation}
    & 1.3          & 71.4              & 2.6       
    & 8.0          & 73.9              & 14.4  
    & 23.5         & 59.2              & 33.6      
    & -            & -                 & -    
    \\
                          ESZSL       \cite{romera2015embarrassingly}
    & 2.4          & 70.1              & 4.6       
    & 5.9          & 77.8              & 11.0  
    & 14.7         & 56.5              & 23.3      
    & -            & -                 & -    
    \\

                     SAE     \cite{kodirov2017semantic}  
    & 0.4           & 80.9              & 0.9        
    & 1.1           & 82.2              & 2.2 
    & 7.8           & 54.0              & 13.6
    & -             & -                 & -    
    \\
    
    \specialrule{.1em}{.00em}{.00em}
\rowcolor[gray]{.9} SDGZSL-ALE      
& 11.1 & 71.8 & \textbf{19.3}
& 21.4 & 88.1 & \textbf{34.4}
& 25.6 & 63.4 & \textbf{36.5}
& 24.8 & 80.0 & \textbf{37.9}    
\\
\midrule
\specialrule{.1em}{.00em}{.00em}
\end{tabular}}
\end{center}
\label{embedding}
\end {table*}
\section{Comparison with Traditional Methods}
To further demonstrate the superiority of the disentangled semantic-consistent representations, we conduct experiments on traditional embeddings methods. Specifically, we use the converged encoder $E_{\psi}$ to process the original visual features. The learned semantic-consistent representations $\vh_s$ substitute the original visual features to learn a compatibility function. We choose the standard embeddings method ALE \cite{akata2015label} as the base method. Table \ref{embedding} shows the performance comparison between our approach and the representative traditional embedding-based approaches. 
It can be seen from the table that traditional embedding-based methods perform poorly on GZSL setting, especially on unseen classes. These traditional embedding-based methods are originally proposed for conventional zero-shot learning setting that only aims to classify test unseen samples over unseen classes.
We argue that the simple embeddings functions cannot draw a clear distinction between the seen class domain and the unseen class domain so that under GZSL setting the unseen class samples tend to be misclassified into seen classes. However, our semantic-consistent representations can alleviate this problem. From the performance table, training with our semantic-consistent representations instead of the original visual features, ALE can boost the performance by a large margin. The improvement verifies the disentangled semantic-consistent representations can help to transfer visual-semantic relationship from seen classes to unseen classes.

\end{document}